\documentclass[10pt,twocolumn]{article}

\usepackage{bbold}
\usepackage{amsmath,amssymb,color,enumerate}
\usepackage{graphicx}
\usepackage{calligra}
\usepackage{mathtools}
\usepackage{algorithm}
\usepackage{algcompatible}

\usepackage{bbm}
\usepackage[margin= 0.8in]{geometry}
\newcommand{\rpm}{\raisebox{.2ex}{$\scriptstyle\pm$}}
\usepackage[round]{natbib}

\usepackage{xcolor}
\usepackage[hidelinks]{hyperref}

\renewcommand\footnotemark{}

\title{Nonlinear System Identification via Tensor Completion }

\author{ \quad \quad \quad Nikos Kargas \footnote{Dedicated to John S. Baras.} \hspace{35px} Nicholas D. Sidiropoulos \\  University of Minnesota  \hspace{5px}  \quad  \quad  University of Virginia \vspace{35 px} \\ 
}
\date{}

\begin{document}
\maketitle

\begin{abstract}
Function approximation from input and output data pairs constitutes a fundamental problem in supervised learning. Deep neural networks are currently the most popular method for learning to mimic the input-output relationship of a general nonlinear system, as they have proven to be very effective in approximating complex highly nonlinear functions. In this work, we show that identifying a general nonlinear function $y = f(x_1,\ldots,x_N)$ from input-output examples can be formulated as a tensor completion problem and  under certain conditions provably correct nonlinear system identification is possible. Specifically, we model the interactions between the $N$ input variables and the scalar output of a system by a single $N$-way tensor, and setup a weighted low-rank tensor completion problem with smoothness regularization which we tackle using a block coordinate descent algorithm. We extend our method to the multi-output setting and the case of partially observed data, which cannot be readily handled by neural networks. Finally, we demonstrate the effectiveness of the approach using several regression tasks including some standard benchmarks and a challenging student grade prediction task.
\end{abstract}

\vspace{10px}
\noindent The problem of identifying a nonlinear function ${y=f(x_1,\ldots,x_N)}$ from input-output examples is of paramount importance in machine learning, dynamical system identification and control, communications, and many other disciplines. In machine learning in particular, most of the supervised learning tasks are nonlinear system identification problems. For example, binary/multiclass classification, where the goal is to predict a discrete variable denoting the class label of each realization, and regression/prediction, where the goal is to predict real or complex valued variables. Algorithmic advancements, availability of vast amounts of data and increasing computational power have led to the development of state-of-the-art prediction models with unprecedented success in various domains such as image classification, speech recognition, and language processing. Kernel methods, random forests, neural networks and deep learning are powerful classes of machine learning models that can learn highly nonlinear functions and have been successfully applied in many supervised machine learning tasks~\cite{HaTiFri2001}. Each of the aforementioned methods can be well suited for a particular problem, but may perform badly for another. In general it is seldom known in advance which method will perform best for any given problem. 

This paper presents a simple and elegant alternative for nonlinear system identification based on low-rank tensor decomposition. Tensor decomposition is a powerful tool for analyzing multi-way data and has had major successes in applications spanning machine learning, statistics, signal processing and data mining~\citep{SiDeFu2017}. The Canonical Polyadic Decomposition (CPD) model is one of the most popular tensor models mainly due to its simplicity and its uniqueness properties. The CPD model has been applied in various machine learning applications, including recommender systems to model time-evolving relational data~\citep{XiCheHu2010}, community detection and clustering to model user interactions across different networks~\citep{papa2013}, knowledge base completion and link prediction for discovering unobserved subject-object interactions~\citep{la2018} and in latent variable models for parameter identification~\citep{anima2014}. These works deal with relatively low-order tensors; however, high-order tensors also arise in practical scenarios -- e.g., a joint probability mass function of $N$ categorical random variables can be naturally regarded as an $N$-th order tensor and modeled using a CPD model~\citep{KaSiFu2018}. 

In this work, we show that the CPD model offers an appealing solution for modeling and learning a {\it general} nonlinear system using a single high-order tensor. Tensors have been used to model {\em low-order multivariate polynomial} systems: a multivariate polynomial of order $d$ is represented by a tensor of order $d$ -- e.g., a second-order polynomial is represented by a quadratic form involving a single matrix \citep{rendle2010}. However, such an approach requires prior knowledge of polynomial order, and assuming that one deals with a polynomial of a given degree can be highly restrictive in practice.
\begin{figure}[!ht]
\centering%
\includegraphics[width=.95\columnwidth]{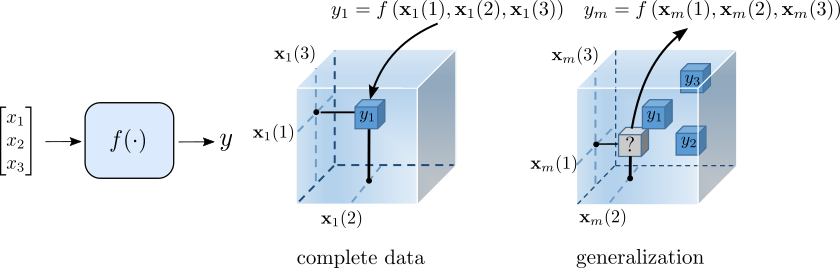}
\caption{Nonlinear system identification as tensor completion.}
\label{fig:completion}
\end{figure}
Instead, what we advocate here is a simple and general approach: a nonlinear system having $N$ discrete inputs and a single output can be naturally represented as an $N$-way tensor where the tuple of input variables $ \left [\mathbf{x}_m(1), \ldots, \mathbf{x}_m(N) \right]$ can be viewed as a cell multi-index and the cell content is the response of the system $y_m$. Given a new data point, the corresponding tensor cell is queried and the output is used as the predictor. 
Note  that only a small fraction of the tensor entries are observed during training, and we are ultimately interested in answering queries for unobserved data points (Figure~\ref{fig:completion}). This motivates the use of low-rank tensor models as a tool for capturing interactions between the predictors and imputing the missing data. 

Both experimental~\citep{tomasi2005} and theoretical studies~\citep{krishnamurthy2013,jain2014,SoDe2019} have shown that exact tensor completion from limited samples is possible under certain conditions. The implication of our simple but profound modeling idea is very compelling, since:
\begin{itemize}
\item The CPD can model any nonlinearity (even of $\infty$ order) for high-enough rank -- because every tensor admits a CPD of bounded rank; see~\citep{SiDeFu2017} and references therein. Even for low ranks, it can model highly nonlinear operators such as products or sums of the signs of the input variables.  

\item Provably correct nonlinear system identification is possible from limited samples. If the associated tensor describing the nonlinear operator is low rank, then it can be fully identified. 

\item In practice, tensors corresponding to real-world systems may not be low-rank; nevertheless even if a system is not exactly low rank, our approach will identify the principal components of the unknown nonlinear mapping, in a sense that will be clarified in the sequel.

\end{itemize}

Even though tensor recovery can be guaranteed under a low-rank assumption, tensor decomposition can often benefit from additional knowledge regarding the application  by incorporating constraints such as non-negativity, sparsity or smoothness~\citep{SiDeFu2017}. In our present context, smoothness is a desirable property for applications where we expect that small perturbations in the input will most probably cause small changes in the output of the system. Therefore, we propose augmenting the CPD tensor completion problem with smoothness regularization on the ordinal latent factors.

\textbf{Contributions}:
We model a general nonlinear system using a single high-order tensor admitting a CPD model.  Specifically, we formulate the problem as a smooth tensor decomposition problem with missing data. Although our method is naturally suited to handle discrete features, it can also be used for continuous valued features~\citep{KaSi2019} and be enhanced using ensemble techniques. Additionally, leveraging the structure of the CPD model, we propose a simple yet effective approach to handle randomly missing input variables. Finally, we discuss how the approach can be extended to vector valued function prediction. The proposed approach requires little parameter tuning, and can model complex nonlinear functions. We propose an easy to implement Block Coordinate Descent (BCD) algorithm and demonstrate the performance in UCI machine learning datasets against competitive baselines as well as a challenging grade prediction task, using real student grade data.

\section{Notation and Background}
We use the symbols $x$, $\mathbf{x}$, $\mathbf{X}$, $\mathcal{X}$ for scalars, vectors, matrices, and tensors respectively. We use the notation  $\mathbf{x}(n)$, $\mathbf{X}(:,n)$, $\mathcal{X}(:,:,n)$ to refer to a particular element of a vector, a column of a matrix and a slab of a tensor. Symbols $\circ$,  $\otimes$, $\circledast$, $\odot$  denote the outer, Kronecker, Hadamard and Khatri-Rao (column-wise Kronecker) product respectively. 

An $N$-way tensor $\mathcal{X} \in \mathbb{R}^{I_1 \times I_2 \times \cdots \times I_N}$ is a multi-dimensional array whose entries are indexed by $N$ coordinates. A polyadic decomposition expresses $\mathcal{X}$ as a sum of rank-$1$ components $\mathcal{X} = \sum_{f=1}^F\mathbf{a}^{1}_f \circ \mathbf{a}^{2}_f \circ \cdots  \circ \mathbf{a}^{N}_f$, where $\mathbf{a}^{n}_f \in \mathbb{R}^{I_n}$. If the number of rank-$1$ components is minimal then the decomposition is called the CPD of $\mathcal{X}$ and $F$ is called the rank of $\mathcal{X}$~\citep{SiDeFu2017}. By defining factor matrices $\mathbf{A}_n = [ \mathbf{a}^{n}_1 \cdots  \mathbf{a}^{n}_F ] \in \mathbb{R}^{I_n \times F}$,  the elements of the tensor $\mathcal{X}$ can be expressed as
\begin{equation}
\mathcal{X}(i_1,\ldots, i_N) = \sum_{f=1}^F \prod_{n=1}^N \mathbf{A}_n(i_n,f).
\label{eq:elementwise}
\end{equation} 
We adopt the common notation ${\mathcal{X} = [\![\mathbf{A}_1,\ldots,\mathbf{A}_N]\!]_F}$ to denote the tensor synthesized from the CPD model using these factors. The mode-$n$ fibers of a tensor are the vectors obtained by fixing all the indices except for the $n$-th index. We can represent tensor $\mathcal{X}$ using a matrix ${\mathcal{X}^{(n)} \in \mathbb{R}^{I_1\cdots I_{n-1} I_{n+1}\cdots I_N \times I_n}}$ called mode-$n$ matricization obtained by arranging the mode-$n$ fibers of the tensor as columns of the resulting matrix
\begin{equation}
\mathcal{X}^{(n)} = \bigl ( \odot_{k\neq n} \mathbf{A}_k \bigr ) \mathbf{A}_n^T,
\end{equation}
where ${\underset{k \neq n}{\odot}\mathbf{A}_k = \mathbf{A}_N \odot \cdots \odot \mathbf{A}_{n+1} \odot \mathbf{A}_{n-1} \odot \cdots \odot \mathbf{A}_1}.$ The $n$-mode product of a tensor $\mathcal{X} \in \mathbb{R}^{I_1\times I_2 \cdots \times I_N}$ with a matrix $\mathbf{U} \in \mathbb{R}^{J \times I_n}$ is denoted by 
$\mathcal{X} \times_n \mathbf{U}$ and an entry of the resulting tensor is given by
\begin{equation}
\begin{aligned}
(\mathcal{X} \times_n \mathbf{U})(i_1,\ldots,i_{n-1},j,i_{n+1},\ldots,i_N) \\ = \sum_{i_n} \mathcal{X}(i_1,\ldots,i_N)\mathbf{U}(j,i_n).
\end{aligned}
\end{equation}
Furthermore, assuming that a tensor $\mathcal{X}$ admits a CPD with rank $F$, the $n$-mode product can be expressed as 
\begin{equation}
\begin{aligned}
&  [ \! [\mathbf{A}_1, \ldots,\mathbf{A}_N] \! ]_F \times_n \mathbf{U} \\ 
& = [\![ \mathbf{A}_1,\ldots,\mathbf{A}_{n-1},\mathbf{U}\mathbf{A}_{n},\mathbf{A}_{n+1}\ldots,\mathbf{A}_N]\!]_F.
\end{aligned}
\end{equation}

CPD is a powerful tool for data analysis mainly due to its uniqueness properties. For a tensor $\mathcal{X}$ of rank $F$, we say that a decomposition $\mathcal{X} = [\![ \mathbf{A}_1, \ldots, \mathbf{A}_N ]\!]_F$ is  unique if the factors are unique up to a common permutation and  scaling / counter-scaling of columns. Specifically, if there exists another decomposition ${ \mathcal{X} = [\![ \widehat{\mathbf{A}}_1, \ldots, \widehat{\mathbf{A}}_N ]\!]_F}$, then, there exists a permutation matrix $\boldsymbol{\Pi}$ and diagonal scaling matrices $\boldsymbol{\Lambda}_1,\ldots, \boldsymbol{\Lambda}_N$  such that $ \widehat{\mathbf{A}}_n = \mathbf{A}_n \boldsymbol{\Pi} \boldsymbol{\Lambda}_n, \forall n\in[N] $ and $\boldsymbol{\Lambda}_1 \cdots \boldsymbol{\Lambda}_N = \mathbf{I}$.
Tensor decomposition is unique under mild rank conditions; see \citep{SiDeFu2017} and references therein. In our context, uniqueness is a desirable property since it is necessary for model interpretability. 

\begin{figure}[!t]
\centering%
\includegraphics[width=.95\columnwidth]{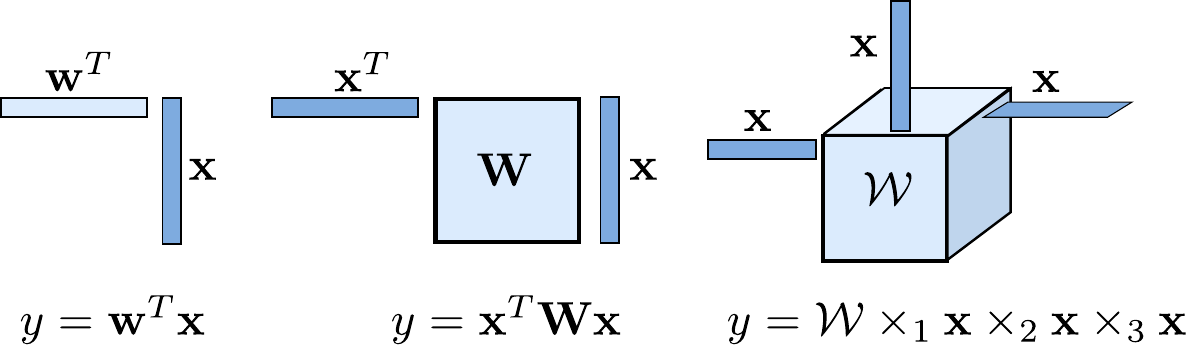}
\caption{Traditional approaches.}
\label{fig:prior}
\end{figure}

\section{Related Work}
Tensors have been mostly used to model low-order multivariate polynomial systems. A multivariate polynomial of degree (order) $d$ can be represented by a tensor of order $d$. For example, a second-order polynomial is represented by a quadratic form involving a single matrix i.e., ${f(\mathbf{x}) = \mathbf{x}^T \mathbf{W} \mathbf{x}}$ while a third order polynomial is represented using a $3$-way tensor i.e., ${ f(\mathbf{x}) = \mathcal{W} \times_1 \mathbf{x} \times_2 \mathbf{x} \times \mathbf{x}_3 } $ (Figure~\ref{fig:prior}). The number of parameters grows exponentially with the order of the approximation making this approach computationally demanding. One way to reduce the number of parameters is to assume that the coefficient tensor is low-rank.

Polynomial Networks (PN) and Factorization Machines (FM) utilize mainly third-order CPD models in order to parameterize the polynomial coefficients with applications in recommender systems and link prediction~\citep{rendle2010,blondel2016b,blondel2016a}. Such approaches require prior knowledge of polynomial order, and assuming that one deals with a polynomial of a given degree can be restrictive. Additionally, even when dealing with the simplest possible approximation model which is rank-$1$, the number of parameters grows linearly with $d$ meaning that this approach cannot model high-degree polynomial functions. Similarly, another tensor model, known as the Tucker model, has  
also been used for parametrization of polynomial functions in a chemogenomics data prediction task~\citep{perros2017}. Finally, a tensor train model~\citep{oseledets2011} has also been used in multivariate polynomial regression~\citep{novikov2016}. Unlike CPD, the model parameters of Tucker and tensor train models are not identifiable. 

Output-only (`blind') identification of {\em linear} systems has also been considered from a tensor point of view. Specifically, identification of Finite Impulse Response (FIR) systems \textit{using only output examples} has been shown in~\citep{bousse2017,Eeghem2018}. 

Our work is radically different  from existing approaches. We model a general nonlinear system using a single tensor of order equal to the number of inputs and propose using a high-order tensor completion approach for system identification. One of the earliest applications of tensor decomposition with smooth latent factors has been fluorescence data analysis~\citep{Bro1998,FuHuMaSi}. Recently, it is has been mostly proposed in the area of image processing. Specifically, CPD and Tucker models with smoothness constraints or regularization have been used for the recovery of incomplete 3- and 4-dimensional image data~\citep{YoZhaCi,ImaHaya}. To the best of our knowledge, tensor completion (with or without smooth latent factors) has not been considered yet as a tool for general nonlinear system identification.

\section{Proposed Approach}

\subsection{Canonical System Identification (CSID)}

We are given a training dataset of $M$ input-output pairs $D = \{( \mathbf{x}_1, y_1),(\mathbf{x}_2, y_2),\ldots, (\mathbf{x}_M, y_M) \}$. Let us assume that all predictors are discrete and take values from a common alphabet $ \mathcal{I} = \{1,\ldots, I\}$. The scalar output $y_m$ is a nonlinear function of the input $\mathbf{x}_m$ distorted by some unknown noise $y_m = f \left (\mathbf{x}_m(1),\ldots, \mathbf{x}_m(N) \right ) + \epsilon_m.$
The nonlinear function $f: \{1,\ldots,I\}^N \rightarrow \mathbb{R}$ can be modeled as an $N$-way tensor $\mathcal{X}$ where each input vector $ \left [\mathbf{x}_m(1), \ldots, \mathbf{x}_m(N) \right]$ can be viewed as a cell multi-index and the cell content is the estimated response of the system $\widehat{y}_m$. We are interested in building a model that minimizes the Mean Square Error (MSE) between the model predictions and the actual response.
However, it is evident that it is impossible to infer the response of unobserved data without any assumptions on $\mathcal{X}$. To alleviate this problem we aim for the principal components of the nonlinear operator by  minimizing the tensor rank. Assuming a low-rank CPD model, the problem of finding the rank-$F$ approximation which best fits our data can be formulated as 
\begin{equation} 
\begin{aligned}
\min_{ \mathcal{X}, \{ \mathbf{A}_n\}_{n=1}^N}  \;  &  \frac{1}{M} \sum_{m=1}^M  \left ( y_m -  \mathcal{X}(\mathbf{x}_m(1),\ldots, \mathbf{x}_m(N)) \right )^2  \\
 & \quad  \quad \quad \quad \quad + \sum_{n=1}^N \rho \| \mathbf{A}_n \|_F^2  \\
\text{s.t. \quad} & \mathcal{X} = \sum_{f=1}^F \mathbf{A}_1(:,f) \odot \cdots  \odot \mathbf{A}_N(:,f),
\end{aligned}
\label{eq:model1}
\end{equation} 
where $\rho$ is a regularization parameter. It is convenient to express the problem in the following equivalent form 
\begin{equation} 
\begin{aligned}
\min_{\mathbf{\mathcal{X}}, \{ \mathbf{A}_n\}_{n=1}^N} &   \frac{1}{M} \left \| \sqrt{\mathcal{W}} \circledast  ( \mathcal{Y} - \mathcal{X} ) \right \|_F^2  
 + \sum_{n=1}^N \rho \| \mathbf{A}_n \|_F^2  \\
\text{subject to \quad} & \mathcal{X} = \sum_{f=1}^F \mathbf{A}_1(:,f) \odot \cdots  \odot \mathbf{A}_N(:,f),
\end{aligned}
\label{eq:model2}
\end{equation} 
where $\mathcal{W}$ is a tensor containing the number of times a particular data point $\mathbf{x} = [i_1,\ldots,i_n]^T$ appears in the dataset and $\mathcal{Y}$ is a tensor containing the mean response of the corresponding data points. The equivalence between Problems~\eqref{eq:model1},~\eqref{eq:model2} is straightforward
\begin{equation*} 
\small
\begin{aligned}
& \min  \sum_{m=1}^M  \Bigl ( y_m - \mathcal{X}(x_m(1), \ldots,  x_m(N)) \Bigr )^2  \Leftrightarrow  \\
& \min \sum_{i_1,\ldots,i_N} \sum_{m' \in S_{i_1,\ldots,i_N}} \bigl ( y_m - \mathcal{{X}}(i_1, \ldots,i_N) \bigr )^2   \Leftrightarrow
\\
&  \min \sum_{i_1,\ldots,i_N}  \sum_{m' \in S_{i_1,\ldots,i_N}}  \bigl ( y_m - \mathcal{Y}(i_1, \ldots,i_N) \\
 & \quad  \quad  \quad  \quad  \quad  + \mathcal{Y}(i_1,\ldots,i_N) - \mathcal{X}(i_1, \ldots,i_N) \bigr )^2 
 \Leftrightarrow \\ &
\min  \sum_{i_1,\ldots,i_N} \mathcal{W}(i_1,\ldots,i_N)
 \bigl ( \mathcal{Y}(i_1,\ldots,i_N)  -  \mathcal{X}(i_1,\ldots, i_N)\bigr)^2.
\end{aligned}
\end{equation*} 
The set $S_{i_1,\ldots,i_N}$ contains the indices the data point $[i_1,\ldots,i_N]^T$ appears in the dataset. Oftentimes, datasets contain both categorical and ordinal predictors, the later, being either discrete or continuous. In the presence of ordinal predictors a desirable property of a regression model is having smooth prediction surfaces i.e., small variations in the input will cause small changes in the output. As an example consider the task of estimating students' grades in future courses based on their grades in past courses, an important topic in educational data mining as it can facilitate the creation of personalized degree paths which will potentially lead to timely graduation~\citep{PoKa2016}. The predictors correspond to the grades in $N$ past courses that a student has received and the the predicted response is the student's grade in a future course. We are interested in building a model that maps an $N$-dimensional discrete feature vector to the output response. Adding a smoothness constraint or regularization will guarantee that the model will produce similar outputs for two students that differ slightly in their past grades as they are likely to perform similarly in the future.
Therefore, we propose augmenting the CPD tensor completion problem with smoothness regularization:
\begin{equation} 
\begin{aligned}
\!\!\! \min_{ \mathcal{X}, \{ \mathbf{A}_n \}_{n=1}^N}  \;  &  \frac{1}{M} \left \| \sqrt{\mathcal{W}} \circledast  ( \mathcal{Y} - \mathcal{X} ) \right \|_F^2  
 + \sum_{n=1}^N \rho \| \mathbf{A}_n \|_F^2 \\ 
 & \quad \quad \quad \quad  + \sum_{n=1}^N \mu_n \| \mathbf{T}_n \mathbf{A}_n \|_F^2 \\
\text{s.t. \quad} & \mathcal{X} = \sum_{f=1}^F \mathbf{A}_1(:,f) \odot \cdots  \odot \mathbf{A}_N(:,f),
\end{aligned}
\label{eq:main_problem}
\end{equation} 
where the matrix $\mathbf{T}_n $ is a smoothness promoting matrix typically defined as $ \mathbf{T}_n \in \mathbb{R}^{(I_n -1) \times I_n}$ with $\mathbf{T}_n(i,i) = 1$ and $\mathbf{T}_n(i,i+1) = -1$ or $ \mathbf{T}_n \in \mathbb{R}^{ (I_n -2)  \times I_n}$ with $\mathbf{T}_n(i,i) = -1$, $\mathbf{T}_n(i,i+1) = 2 $ and $\mathbf{T}_n(i,i+2) = -1 $. We set $\mu_n = 0$ for categorical predictors and $\mu_n > 0$ otherwise. Penalizing the difference of consecutive row elements of a factor $\mathbf{A}_n$ guarantees that varying the $n$-th dimension and keeping the remaining fixed will have a small impact on the predicted response. Another appealing feature of the proposed smoothness regularization is that it can potentially measure feature importance. Note that the effect a  variable will have in the prediction is minimized if each column of the corresponding factor is a constant number. Irrelevant features are more likely to have factors that vary slightly. On the contrary, factors associated with predictive features will have more variations and induce a larger penalty cost.

\textbf{Remark}: CPD can model any nonlinear operator for high-enough rank, but even for low ranks, it can model highly nonlinear operators such as 
\begin{equation*}
\begin{aligned}
f_1(x_1,\ldots,x_N) = \prod_{n=1}^N {\rm sign}(x_n), \\  f_2(x_1,\ldots,x_N) = \sum_{n=1}^N {\rm sign}(x_n).
\end{aligned}
\end{equation*}
Comparing these equations with Equation~\eqref{eq:elementwise} we can verify that the former corresponds to a rank-$1$ CPD model, while the later to rank $N$. 

\subsection{Tensor Completion: Identifiability}

In this section we briefly review existing probabilistic and deterministic theoretical results on tensor recovery from a few samples. This is important because, using our approach of casting system identification as tensor completion, the results below directly yield new results on nonlinear multivariate system identification - even for systems of unbounded nonlinearity degree. Recovering a tensor from samples depends mainly on how the samples $(X_1,\ldots,X_N)$ are generated -- randomly or systematically, and if randomly from what distribution -- as well as the operational ${\bf A}_n$. Practical experience suggests that the generic sample complexity for randomly drawn point samples is proportional to the degrees of freedom $O(FNI)$ in the model. This has been proven for randomly drawn {\em linear} (generalized, aggregated) samples, but not yet for point samples~\citep{bousse2018}.

An adaptive sampling method with an estimation algorithm has been proposed by~\citep{krishnamurthy2013} that provably recovers an $N$-th order rank-$F$ tensor using $O(IF^{N-0,5} \mu^{N-1} N \log F)$ samples where $\mu$ is a coherence bound on the factor matrices. Later, \citep{jain2014} proposed a method that can recover an $N$-th order rank-$F$ tensor with orthogonal factor matrices and random sampling using $O(I^{N/2} \mu^6 F^5 \log^4I)$ samples. A necessary condition for these methods is that the rank needs to be less than the maximum outer dimension although a CPD model can be unique even if rank exceeds this bound. Probabilistic results on tensor completion have also been proven for incomplete tensors that have low mode-$n$ ranks under certain incoherence conditions, relying on minimization of the sum of nuclear norms of the tensor unfoldings~\citep{gandy2011}. For $F < I$, the result in~\citep{Yuan2016} can be used to show that for uniform random point samples, the sample complexity for our low-rank model is $O(\sqrt{F} I^{N/2} \log I)$. 

Deterministic conditions based on a specific sampling strategy, namely fiber sampling, have been given by \citep{SoDe2019}. Necessary and sufficient conditions are provided which are dependent on the sampling pattern, assuming  that the rank is low enough. The authors propose an eigenvalue decomposition algorithm and demonstrate exact recovery of low-rank incomplete tensors even when, less than one percent of the tensor entries are available. The authors have also extended the results in the case of not fully observed fibers. Finally, several {\em regular} sampling strategies are investigated and generic identifiability conditions are provided by~\citep{kanatsoulis2019}.

\subsection{Algorithm}
\label{sec:alg}

The work-horse of tensor decomposition is the so-called Alternating Least Squares (ALS) algorithm. ALS is a special type of BCD which offers two distinct advantages: monotonic decrease of the cost function, and no need for parameter tuning. In this section, we propose an ALS approach to tackle Problem~\ref{eq:main_problem}. 

Tensors $\mathcal{W},\mathcal{Y}$ despite being high-dimensional, are in general very sparse and optimized sparse tensor formats can offer huge memory and computational savings~\citep{smith2015tensor}. The idea of ALS is that we cyclically update variables $\{\mathbf{A}_n\}_{n=1}^N$ while fixing the remaining variables at their last updated values. Assume that we fix estimates $\mathbf{A}_n$, $ \forall n \in [N] \setminus \{k\}$ we need to solve the following optimization problem
\begin{equation}
\begin{aligned}
& \min_{\mathbf{A}_k}  \bigl \| { \widehat{\mathcal{W}}^{(k)}} \circledast  \mathcal{Y}^{(k)} -  {\widehat{\mathcal{W}}^{(k)}} \circledast  (\mathbf{Q}_k \mathbf{A}_k^T)  \bigr \|_F^2  \\ &  \quad \quad \quad \quad \quad +  \rho \| \mathbf{A}_k \|_F^2 + \mu_k \| \mathbf{T}_k \mathbf{A}_k \|_F^2,
\end{aligned}
\end{equation}
where $\mathbf{Q}_k = (\odot_{n\neq k} \mathbf{A}_n)$,  $\widehat{\mathcal{W}} = \sqrt{\mathcal{W}}$ with the square root computed element-wise. Equivalently, we have
\begin{equation}
\begin{aligned}
& \min_{\mathbf{A}_k}  \sum_{i_k=1}^{I_k} \Bigl \| {\rm{diag}} ( \widehat{\mathbf{w}}^k_{i_k} )  \left ( \mathbf{y}^k_{i_k} -  \mathbf{Q}_{k} \mathbf{a}^k_i \right) \Bigr \|_2^2 \\
& \quad \quad \quad \quad \quad +  \rho \| \mathbf{A}_n \|_F^2 + \mu_k \| \mathbf{T}_k \mathbf{A}_n \|_F^2,
\end{aligned}
\label{eq:convex}
\end{equation}
where $\widehat{\mathbf{w}}^k_{i_k} = { \widehat{\mathcal{W}}^{(k)}(:, i_k)}$, $\mathbf{y}^k_{i_k} = {\mathcal{Y}^{(k)}(:, i_k)}$ and $\mathbf{a}^k_i = \mathbf{A}_k(i_k,:)^T$. Note that we do not need to instantiate $\mathbf{Q}_{k}$ because only the non-zero elements of the sparse vector $\widehat{\mathbf{w}}^k_{i_k}$ contribute to the cost function. The non-zero elements of $\widehat{\mathbf{w}}^k_{i_k}$ correspond to the observed data points for which the $k$-th variable takes the value $i_k$ and therefore we need to compute the corresponding rows of the Khatri-Rao product. Problem~\ref{eq:convex} can be optimally solved by finding the solution to a set of linear equations obtained after setting the gradient to zero e.g., using the conjugate Gradient descent algorithm~\citep{bertsekas1997}. Simpler updates can be obtained by fixing all variables except for a single row of the factor $\mathbf{A}_k$. Let us fix every parameter except for the $i_k$-th row of $\mathbf{A}_k$ 
\begin{equation}
\begin{aligned}
& \min_{\mathbf{a}^k_i} \frac{1}{M} \bigl\|  {\rm{diag}} ( \widehat{\mathbf{w}}^k_{i_k} )   (\mathbf{y}^k_i  -  \mathbf{Q}_k \mathbf{a}^k_i \bigr \|_2^2 + \rho \| \mathbf{a}^k_i \|_2^2 
\\ 
& \quad \quad \quad +  \mu_k \left \|\mathbf{a}^k_{i-1} -  \mathbf{a}^k_i \right \|_2^2 +  \mu_k \left \| \mathbf{a}^k_{i+1} - \mathbf{a}^k_i \right \|_2^2,
\end{aligned} 
\end{equation}
The solution for $\mathbf{a}^k_i$ is given by 
\begin{equation}
\begin{aligned}
& \mathbf{a}^k_i = (\mathbf{Q}_k^T {\rm{diag}}(\mathbf{w}_i)^2 \mathbf{Q}_k + (\rho +  2\mu_k) \mathbf{I} )^{-1} 
\\
& \quad \quad \quad ( \mathbf{Q}_k^T {\rm{diag}}(\mathbf{w}_i)^2 \mathbf{y}^k_i   -\mu_k( \mathbf{a}^k_{i-1} + \mathbf{a}^k_{i+1} ))
\end{aligned}
\label{eq:ls}
\end{equation} 
which results in very lightweight row-wise updates. BCD algorithms usually offer faster convergence in terms of the cost function compared to stochastic algorithms for small or moderate size problems. For large-scale problems on the other hand, Stochastic Gradient Descent (SGD) can attain moderate solution accuracy faster than BCD. The merits of both alternating optimization and stochastic optimization can be combined by considering block-stochastic updates~\citep{XuYin}. In this work, we propose an easy to implement ALS algorithm as our main goal is to present a fresh perspective on the nonlinear identification problem through low-rank tensor completion. Further algorithmic developments are underway, but beyond the scope of this first submission. Next, we show how the proposed approach can be extended to handle partially observed and multi-output regression tasks.

\begin{table*}[!ht]
\caption{ Comparison of RMSE performance of different models on UCI datasets without missing data.}
\centering
\resizebox{.95\textwidth}{!}{
\begin{tabular}{|c | c | c | c | c | c | c |}
\hline 
Dataset    & RR  &  SVR (RBF) &  SVR (polynomial) & DT & MLP (5 Layer)  & CSID \\
\hline
Energy Eff. (1)  & $ 2.91 \rpm 0.17 $ &  $ 2.68 \rpm  0.17  $ & $  4.09 \rpm 0.49 $  & $  0.56 \rpm  0.03 $ & $  \mathbf{ 0.48 \rpm  0.06~[50] } $ & $ \mathbf{0.39 \rpm 0.05}  $ \\ 
Energy Eff. (2) & $  3.09 \rpm  0.19 $ &  $  3.03 \rpm 0.21 $ & $ 4.14 \rpm  0.44  $  & $ 1.86 \rpm  0.19 $ & $ \mathbf{ 0.97 \rpm 0.14~[50] }$  & $ \mathbf{ 0.57 \rpm  0.09} $ \\ 
\hline
C. Comp. Strength & $ 10.47 \rpm  0.42$ &  $ 9.72  \rpm 0.38   $ & $  11.30 \rpm 0.36  $  & $  6.57 \rpm  0.82 $ & $  \mathbf{ 4.92
 \rpm  0.63~[50]} $ & $  \mathbf{ 4.67\rpm 0.50} $ \\ 
\hline
SkillCraft Master Table & $ 1.68  \rpm 1.61  $ &  $  \mathbf{ 0.99  \rpm  0.03  }$ & $ 1.22  \rpm  0.05  $  & $ 1.03 \rpm  0.04  $ & $  1.00 \rpm  0.03~[10]  $ & $\mathbf{  0.91 \rpm 0.02} $ \\ 
\hline
Abalone & $ 2.25\rpm 0.10  $ &  $ \mathbf{  2.19\rpm   0.08 } $ & $ 3.90\rpm  3.43 $ & $  2.35 \rpm 0.08   $ & $ \mathbf{ 2.09 \rpm 0.09~[10]} $ & $  2.23 \rpm 0.09 $ \\ 
\hline
Wine Quality & $ 0.76 \rpm 0.02 $ &  $ \mathbf{ 0.69 \rpm   0.02}  $ & $ 1.01 \rpm  0.39  $ & $ 0.75 \rpm   0.03  $ & $ 0.72 \rpm 0.02~[10]  $ & $ \mathbf{ 0.70 \rpm 0.02}  $ \\ 
\hline
Parkinsons Tel. (1) & $ 7.51 \rpm  0.11  $ &  $   6.66 \rpm  0.14   $ & $  7.89 \rpm  0.88  $  & $ \mathbf{ 2.40 \rpm  0.26 } $ & $ 3.60 \rpm 0.18 ~[100] $ & $\mathbf{ 1.33 \rpm 0.10  }$ \\ 
Parkinsons Tel. (2) & $  9.75  \rpm 0.15 $ &  $  9.14 \rpm   0.17 $ & $   10.04 \rpm  0.43  $  & $  \mathbf{ 2.60 \rpm   0.38 }$& $  5.01 \rpm  0.19~[100]  $ & $  \mathbf{ 1.79 \rpm 0.17 } $ \\ 
\hline
C. Cycle Power Plant & $   5.51 \rpm  0.09 $ &  $  4.13 \rpm   0.09 $ & $ 8.00  \rpm  0.19 $  & $   \mathbf{ 3.98 \rpm  0.13  }   $ & $  4.06 \rpm 0.11~[50] $ & $\mathbf{ 3.76 \rpm 0.15 } $ \\ 
\hline
Bike Sharing (1) & $   36.45  \rpm 0.46   $ &  $  32.67 \rpm  0.81 $ & $ 34.93 \rpm 0.97 $ & $   18.89 \rpm   0.36 $ & $ \mathbf{ 14.81 \rpm  0.44~[100] } $ & $ \mathbf{ 15.17 \rpm 0.44 } $ \\ 
Bike Sharing (2) & $  122.65 \rpm   2.87    $ & $   113.18 \rpm  1.73   $ & $  117.25 \rpm  2.01  $  & $  42.06 \rpm  2.06  $ & $\mathbf{ 38.69 \rpm 1.24~[100] } $ &  $ \mathbf{ 36.93 \rpm 1.19 } $ \\ 
\hline
Phys. Prop. &  $ 5.19 \rpm 0.03   $ & $ 4.91 \rpm 1.26  $ & $  6.49 \rpm 1.15  $  & $ 4.40 \rpm 0.04    $ & $  \mathbf{4.20 \rpm  0.05~[100] }$ & $   \mathbf{ 4.21 \rpm  0.04} $ \\ 
\hline
\end{tabular}
}
\label{table:full_data}
\end{table*}

\begin{table*}[!ht]
\caption{ Comparison of RMSE performance of different models on UCI datasets with $30\%$ missing data.}
\centering
\resizebox{.95\textwidth}{!}{
\begin{tabular}{|c | c | c | c | c | c | c | c |}
\hline 
Dataset    & RR  &  SVR (RBF) &  SVR (polynomial) & DT & MLP (5 Layer) & CSID \\
\hline
Energy Eff. (1)  & $ 3.01 \rpm 0.15     $ &  $ 3.38 \rpm  0.27     $ & $   6.88 \rpm 0.63        $  & $  2.57 \rpm   0.49   $ & $ \mathbf{ 2.49 \rpm 0.48~[10] } $ & $  \mathbf{ 2.17  \rpm 0.25} $ \\ 
Energy Eff.  (2) & $   3.26  \rpm  0.16     $ &  $  3.57 \rpm  0.30     $ & $ 6.65 \rpm  0.48        $  & $  \mathbf{ 2.64 \rpm  0.28 }     $ & $ 3.02 \rpm   0.36 ~[10]  $ & $  \mathbf{2.48  \rpm 0.22 } $ \\ 
\hline
C. Comp. Strength  & $  10.33 \rpm  0.61       $ &  $ 11.39 \rpm   0.48      $ & $ 13.16  \rpm  1.17    $  & $  \mathbf{ 9.90  \rpm  1.05 } $ & $  10.01\rpm  0.54 ~[10]  $  & $  \mathbf{ 9.69 \rpm 0.79  }$ \\ 
\hline
SkillCraft Master Table  & $    1.79 \rpm   1.63   $ &  $  \mathbf{  1.05  \rpm   0.03  } $ & $   1.61 \rpm 0.33      $  & $    1.08 \rpm  0.03         $ & $  1.10 \rpm 0.04~[10]  $  & $  \mathbf{ 1.05  \rpm 0.01 } $ \\ 
\hline
Abalone & $     \mathbf{ 2.27 \rpm  0.07 }$ &  $   2.31  \rpm   0.08  $ & $   3.12 \rpm  0.79  $ & $  2.42 \rpm  0.07    $ & $  \mathbf{ 2.28 \rpm 0.07 ~[10] }$ & $   2.40  \rpm 0.13  $ \\ 
\hline
Wine Quality & $     0.76 \rpm  0.02   $ &  $ \mathbf{ 0.73 \rpm 0.02    } $ & $ 0.93  \rpm  0.21   $ & $  0.78 \rpm   0.02   $ & $  \mathbf{0.76 \rpm 0.03 ~[10]}      $ & $ 0.78 \rpm 0.02 $ \\ 
\hline
Parkinsons Tel. (1) & $ 7.52  \rpm  0.11     $ &  $ 6.91  \rpm 0.13    $ & $  8.12  \rpm   0.11 $  & $   \mathbf{ 3.10 \rpm 0.22 } $ & $ 5.90 \rpm 0.28 ~[10]  $ & $  \mathbf{4.98 \rpm 0.12} $  \\ 
Parkinsons Tel. (2) & $   9.76 \rpm 0.18    $ &  $   9.38  \rpm 0.21  $ & $ 10.68  \rpm  0.23       $  & $ \mathbf{  3.59 \rpm   0.81}   $ &  $   7.67  \rpm 0.18~[10] $ & $ \mathbf{6.58 \rpm 0.18 } $ \\ 
\hline
C. Cycle Power Plant & $  5.51  \rpm   0.09$ &  $     6.16 \rpm  0.15  $ & $  10.45  \rpm 0.31       $ & $  \mathbf{ 5.29 \rpm 0.36 }   $ & $   5.33  \rpm  0.07~[50] $ & $ \mathbf{ 5.04 \rpm 0.12} $ \\ 
\hline
Bike Sharing (1) & $  37.40 \rpm 0.52 $ &  $ 35.50  \rpm 0.31  $ & $  36.85 \rpm  0.38 $ & $  25.41 \rpm 1.5 $ & $    \mathbf{ 21.51 \rpm 0.83  \rpm~[50] }   $ & $\mathbf{ 23.89 \rpm  0.19 }$ \\ 
Bike Sharing (2) & $ 123.81 \rpm  1.26 $ & $ 127.06 \rpm 1.55   $ & $  130.20 \rpm 1.13  $  & $ \mathbf{ 71.93  \rpm 1.18 }$&  $    \mathbf{ 64.03 \rpm  1.66 ~[50] } $ & $ 75.65  \rpm 1.51 $ \\ 
\hline
Phys. Prop. &  $  5.18 \rpm 0.02     $ & $  7.53 \rpm 0.67  $ & $ 7.87 \rpm 0.83  $   & $ 5.08 \rpm 0.03     $ & $  \mathbf{4.99  \rpm 0.09~[100]} $& $  \mathbf{ 4.70 \rpm  0.03 }  $  \\ 
\hline
\end{tabular}
}
\label{table:missing}
\end{table*}

\begin{table*}[!ht]
\caption{Comparison of RMSE performance of different models on multi-output regression.}
\centering
\resizebox{.95\textwidth}{!}{
\begin{tabular}{|c | c | c | c | c | c | c | c |}
\hline 
Dataset    & RR  &   MLP (1 Layer) & MLP (3 Layer) & MLP (5 Layer) & DT & CSID \\
\hline
En. Eff.  (2) & $2.70 \rpm  0.19  $ &  $  2.82 \rpm  0.08 ~[50]  $ & $  2.73 \rpm 0.11[100]  $  & $   2.67 \rpm  0.11[10]   $ & $ \mathbf{ 2.19 \rpm  0.19} $ & $ \mathbf{ 2.01 \rpm 0.14 }$ \\ 
\hline
Park. Tel. (2) & $ 12.19 \rpm  0.09  $ &  $   7.59 \rpm  0.21[250] $ & $   6.54 \rpm  0.06[250]   $  & $ 6.18 \rpm 0.42[250]  $ & $ \mathbf{ 3.37 \rpm 0.39 }  $ & $ \mathbf{2.85 \rpm 0.22} $   \\ 
\hline
B. Shar. (2) & $  127.75  \rpm 3.32  $ &  $   64.12 \rpm 6.49[250]  $ & $43.60 \rpm 1.95[100] $& $ \mathbf{42.25 \rpm 1.22[100]}$   & $ 46.21 \rpm  1.20 $ & $ \mathbf{ 45.29 \rpm 1.47 } $  \\ 
\hline
\end{tabular}
}
\label{table:multi_outpu}
\end{table*}

\subsection{Missing Data}
It is quite common in general to have observations with missing values for one or more predictors. For example, in the grade prediction task described in the introduction, the predictions for a student rely on the student's performance achieved in previously taken courses. Consider a student-grade matrix $\mathbf{D} \in {\mathbb{R}}^{M \times N}$ where our goal is to predict the $N$-th course. The matrix will be in general sparse since each student enrolls in only few of the available courses, and the selected courses vary from student to student. 

Common approaches for handling missing data include ($1$) removal of observations with any missing values, ($2$) imputing the missing values before training e.g., by replacing them with the mean, median, or the mode, and ($3$) directly handling the imputation by the algorithm. Let $\mathcal{O} = \{o_1,\ldots,o_T\}$ and $\mathcal{M} = \{m_1,\ldots,m_L\}$ denote the indices of the observed and missing entries of a single observation respectively. Instead of ignoring observations with missing entries we aim at computing the expectation of the nonlinear function conditioned on the observed variables i.e., we set
\begin{equation}
\begin{aligned}
 f(\mathbf{x}_{\mathcal{O}})&  =  \mathbb{E}_{\mathbf{x}_{\mathcal{M}} | \mathbf{x}_{\mathcal{O}} } [ f( \mathbf{x}_\mathcal{O}, \mathbf{x}_\mathcal{M}) ] 
\\ & = \sum_{\mathbf{x}_{\mathcal{M} }} {\sf Pr}( \mathbf{x}_{\mathcal{M}} \vert \mathbf{x}_{\mathcal{O}}) f( \mathbf{x}_\mathcal{O}, \mathbf{x}_\mathcal{M}).
\end{aligned}
\end{equation}
Estimating the conditional probability ${\sf Pr}( \mathbf{x}_{\mathcal{M}} \vert \mathbf{x}_{\mathcal{O}})$ is not possible since the number of parameters grows exponentially with the number of missing entries. Given the low-rank structure of the nonlinear function we propose modeling the Probability Mass Function (PMF) using a nonnegative CPD model which is a universal model for PMF estimation~\citep{KaSiFu2018}. For the sake of simplicity, we adopt a simple rank-one joint PMF model estimated via the empirical first-order marginals~\citep{huang2017kullback}. Without loss of generality assume that the first $T$ predictors are known and the remaining missing, then, the expectation can be computed very efficiently
\begin{equation}
\begin{aligned}
& f(\mathbf{x}_{\mathcal{O}})  =   \mathbb{E}_{\mathbf{x}_{\mathcal{M}} | \mathbf{x}_{\mathcal{O}} } [ f( \mathbf{x}_\mathcal{O}, \mathbf{x}_\mathcal{M}) ] \\ 
& = \mathcal{X}(i_1,\ldots,i_T,:,\ldots,:) \times_{T+1} \mathbf{p}_{T+1}\cdots \times_{T+L} \mathbf{p}_{N} \\
 &= \sum_{f=1}^F \prod_{n=1}^T \mathbf{A}_n(i_n,f) \prod_{n=T+1}^N \mathbf{p}_n^T \mathbf{A}_n(:,f).
\end{aligned}
\end{equation}
In this case, we minimize the squared error between the target value and the conditional expectation of the function. The modification can be easily incorporated in the ALS algorithm. Rich dependencies between the variables can also be captured using a higher-order PMF model, but we defer this discussion to follow-up work due to space limitations.

\subsection{Multi-Output Regression}
The proposed framework is quite flexible and can easily be extended to vector-valued functions ${f: \{1,\ldots,I\}^N \rightarrow \mathbb{R}^K}$. When there is no correlation between the output variables of a system, one can build $K$ independent models, one for each output, and then use those models to independently predict each one of the $K$ outputs. However, it is likely that the output values related to the same input are themselves correlated and often a better way is to build a single model capable of predicting simultaneously all $K$ outputs. We can treat each different model as an $N$-way tensor and stack them together to build an $(N+1)$-way tensor. The new tensor model can be described by $N+1$ factors associated with the $N$ predictors and an additional mode of dimension $K$, $\mathcal{X} = [\![\mathbf{A}_1,\ldots,\mathbf{A}_N,\mathbf{V}]\!]_F$. The vector-valued prediction for $[i_1, \ldots, i_N]^T$ is given by
${\mathcal{X}(i_1, \ldots, i_N, j) = \sum_{f=1}^F \mathbf{V}(j,f) \prod_{n=1}^N \mathbf{A}_n(i_n,f)}.
$ In matrix form we have
\begin{equation*}
\mathcal{X}(i_1, \ldots, i_N, :) =   \left( \mathbf{A}_1(i_1,:) \circledast \cdots \circledast  \mathbf{A}_N(i_N,:) \right ) \mathbf{V}^T
\end{equation*}
No modification is needed for the ALS updates.  Depending on the application one may or may not need to apply smoothness regularization on $\mathbf{V}$.

\section{Experiments}
We evaluate the proposed approach in single output regression tasks using several datasets obtained from the UCI machine learning  repository~\citep{UCI}. Our proposed approach is implemented in MATLAB using the Tensor Toolbox~\citep{TTB_Sparse} for tensor operations. We then assess the ability of our model to handle missing predictors by hiding $30\%$ of the data as well as its ability to predict vector valued responses. For each experiment we split the dataset into two sets, $80\%$ used for training and $20\%$ for testing, and run $10$ Monte-Carlo simulations. Finally, we evaluate the performance of our approach in a challenging student grade prediction task  using a real student grade dataset. For each method we tune the hyper-parameters using $5$-fold cross-validation. We compare the performance of the different algorithms in terms of the Root Mean Square Error (RMSE).

\subsection{UCI Datasets}
We used four different machine learning algorithms as baselines, Ridge Regresion (RR), Support Vector Regression (SVR), Decision Tree (DT) and  Multilayer Perceptrons (MLPs) using the implementation of scikit-learn~\citep{scikit2011}. For RR, SVR and MLP we standardize each ordinal feature such that it has zero mean and unit variance. Categorical features are transformed using one-hot encoding. For DT no preprocessing step is required. For our method, we fix the alphabet size to be $I=25$ and use Lloyd-Max scalar quantizer for discretization of continuous predictors. For the MLPs,  we set the number of hidden layers to $1$,$3$ or $5$ and varied the number of nodes per layer $10, 50$, $100$ and $250$. We observed that in most cases the MLP with $5$ hidden layers performed better than the $1$ or $3$ layer MLP and that further increasing the number of layers did not improve the performance. 

Table~\ref{table:full_data} shows the RMSE performance of the different methods when there are no missing predictors on the datasets. The number inside the square brackets denotes the number of nodes for each layer of MLP. We highlight the two best performing methods for each dataset. Our approach performs  similarly or better than best baseline in most of the datasets. Note that both decision trees and our approach rely on discretization of continuous predictors however, adding the smooth regularization plays a significant role in boosting the RMSE performance for our method. 

Next, we evaluate our approach on partially observed datasets. We randomly hide $30\%$ of the full dataset and repeat $10$ Monte-Carlo simulations. Before fitting the data to the baseline algorithms we replace each missing entry of an ordinal predictor with the mean and for each categorical predictor we use the most frequent value (mode). For our algorithm we use a rank-$1$ approximation of the joint PMF tensor estimated from the training data. Table~\ref{table:missing} shows the performance of the different algorithms in this setting. Again, our approach similarly or better than best baseline.

Finally, we test our approach in predicting multi-output responses against RR, DT tree and MLPs.  Table~\ref{table:multi_outpu} contains the results for three datasets. Similarly to the single output setting our approach performs the same or slightly better compared to the baseline methods.

\begin{table*}[!t]
\caption{Comparison of RMSE performance on student grade data.}
\centering
\resizebox{.95\textwidth}{!}{
\begin{tabular}{|c | c  | c|  c  | c|}
\hline 
Dataset & GPA  &  BMF & CSID   \\
\hline
\hline
CSCI-1 	&  $0.52 \rpm 0.02$  &  $\mathbf{0.48 \rpm 0.03}$ &   $\mathbf{0.48 \rpm 0.03}$\\ 
CSCI-2  &  $0.56 \rpm 0.02$  &  $\mathbf{0.55 \rpm 0.02}$ &   $\mathbf{0.55 \rpm 0.03}$\\ 
CSCI-3  &  $\mathbf{0.48 \rpm 0.04}$   &  $\mathbf{0.48 \rpm  0.04}$ &   $\mathbf{0.48 \rpm  0.05}$\\ 
CSCI-4  &  $0.53 \rpm 0.03$  &      $0.52 \rpm 0.04 $ &   $\mathbf{0.51 \rpm 0.03}$\\ 
CSCI-5  &  $0.43 \rpm 0.02$  &     $0.43 \rpm 0.02$ &   $\mathbf{0.42 \rpm 0.02}$\\ 
CSCI-6  &  $0.63 \rpm 0.03$  &  $0.58 \rpm  0.03$ &   $\mathbf{0.57 \rpm 0.03}$\\ 
CSCI-7  &  $0.57 \rpm 0.02$  &  $0.58 \rpm 0.01$ &   $\mathbf{0.56 \rpm 0.02}$\\ 
CSCI-8  &  $0.52 \rpm 0.02$  &  $0.49 \rpm 0.03$ &   $\mathbf{0.47 \rpm 0.02}$\\ 
CSCI-9  &  $0.61 \rpm 0.03$  &  $0.60 \rpm 0.05$ &   $\mathbf{0.57 \rpm 0.03}$\\ 
CSCI-10 &  $0.58 \rpm 0.04$  &  $\mathbf{0.56 \rpm 0.04}$ &   $\mathbf{0.56 \rpm 0.04}$\\ 
\hline
\end{tabular} 
\quad \quad
\begin{tabular}{|c | c|  c  | c|}
\hline 
Dataset & GPA  &  BMF & CSID   \\
\hline
\hline
CSCI-11  &  $0.68 \rpm 0.06 $   &  $\mathbf{0.66 \rpm 0.04 }$ &   $0.67 \rpm 0.03$\\ 
CSCI-12  &  $0.58 \rpm 0.04$   &  $0.51 \rpm 0.04$ &   $\mathbf{0.48 \rpm 0.01}$\\ 
CSCI-13  &  $0.67 \rpm 0.03$   &  $0.55 \rpm 0.05$ &   $\mathbf{0.54 \rpm 0.03}$\\ 
CSCI-14  &  $0.70 \rpm 0.06$   &  $\mathbf{0.62 \rpm  0.03}$ &   $0.65 \rpm 0.07$\\ 
CSCI-15  &  $0.56 \rpm 0.03$   &  $\mathbf{0.56 \rpm 0.06}$ &   $0.57 \rpm 0.03 $\\ 
CSCI-16  &  $0.52 \rpm 0.03$   &  $0.51 \rpm  0.03$ &   $\mathbf{0.50 \rpm 0.02}$\\ 
CSCI-17  &  $0.60 \rpm 0.02$   &  $\mathbf{0.58 \rpm 0.05}$ &   $0.59 \rpm  0.05$ \\
CSCI-18  &  $0.57 \rpm 0.03$   &  $0.56 \rpm 0.05$ &   $\mathbf{0.55 \rpm 0.04}$\\
CSCI-19  &  $0.68 \rpm 0.04$   &  $0.70 \rpm 0.04$ &   $\mathbf{0.61 \rpm  0.04}$\\ 
CSCI-20  &  $0.61 \rpm 0.06$   &  $\mathbf{0.58 \rpm 0.02}$ &   $0.63 \rpm  0.04$\\ 
\hline
\end{tabular}
}
\label{table:grade_prediction}
\end{table*}

\begin{figure}[!t]
\centering
\includegraphics[width=.95\columnwidth]{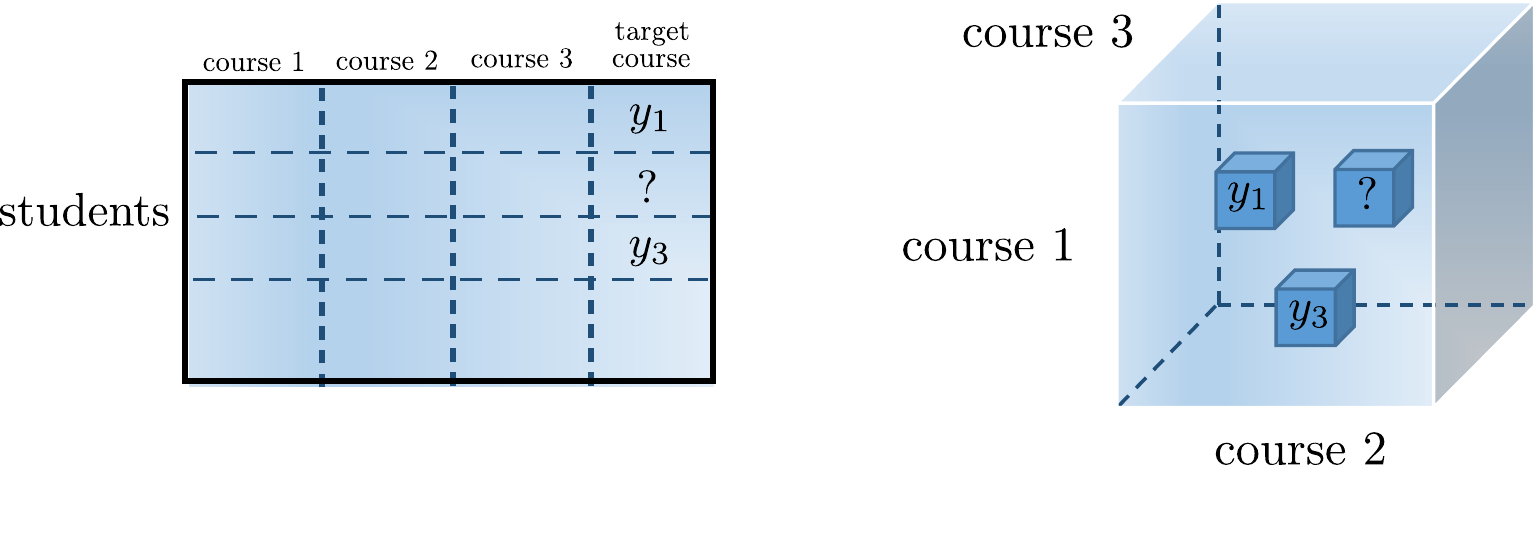}
\caption{Low-rank matrix completion (left)  canonical system identification (right).}
\label{fig:mat_ten}
\end{figure}
\subsection{Grade Prediction Datasets}
Finally we evaluate our method in a student grade prediction task on a real dataset obtained from the CS department of a university. The predictors corespond to the course grades the students have received. Specifically, we used the $20$ most frequent courses to build $20$ independent single output regression tasks each one of them having $34$ predictors. Grades take $11$ discrete values ($A$-$F$) and due to the natural ordering between the different values smoothness regularization was applied on all factors. We used the Grade Point Average (GPA) and Biased Matrix Factorization as our baselines. Low-rank matrix completion is considered a state-of-art method in student grade prediction~\citep{PoKa2016,almutairi2017}. Note that in the matrix case each course is represented by a column while in the proposed tensor approach, each course is represented by a tensor mode (Figure~\ref{fig:mat_ten}). Table~\ref{table:grade_prediction} shows the results for the different algorithms. Our approach outperforms BMF in $11$ tasks, performs the same in $4$ and worse in $5$. 

\section{Conclusion and Future work}
In this paper, we considered the problem of nonlinear system identification. We formulated the problem as a smooth tensor completion problem with missing data and developed a lightweight BCD algorithm to tackle it. We have proposed a simple approach to handle randomly missing data and extended our model to vector valued function approximation. Experiments on several real data regression tasks showcased the effectiveness of the proposed approach. 

\section{Acknowledgements}
The work of the authors was supported in part by NSFIIS-1447788, IIS-1704074.

\small
\bibliography{lib}

\end{document}